\documentclass[sigconf,nonacm]{acmart}
\usepackage{algorithm}
\usepackage{algorithmic}
\AtBeginDocument{%
  }

\setcopyright{acmlicensed}
\copyrightyear{2018}
\acmYear{2018}
\acmDOI{XXXXXXX.XXXXXXX}
\acmConference[Conference acronym 'XX]{Make sure to enter the correct
  conference title from your rights confirmation email}{June 03--05,
  2018}{Woodstock, NY}
\acmISBN{978-1-4503-XXXX-X/2018/06}

\begin{document}

\title{Fast MoE Inference via Predictive Prefetching and Expert Replication}

\author{Ankit Jyothish}
\affiliation{%
  \institution{Iowa State University}
  \city{Ames}
  \country{USA}}
\email{ankitj99@iastate.edu}

\author{Ali Jannesari}
\affiliation{%
  \institution{Iowa State University}
  \city{Ames}
  \country{USA}}
\email{jannesar@iastate.edu}

\author{Aishwarya Sarkar}
\affiliation{%
  \institution{Iowa State University}
  \city{Ames}
  \country{USA}}
\email{asarkar1@iastate.edu}

\author{Joseph Zuber}
\affiliation{%
  \institution{Iowa State University}
  \city{Ames}
  \country{USA}}
\email{zuber@iastate.edu}

\renewcommand{\shortauthors}{Trovato et al.}


\begin{abstract}
The Mixture of Experts (MoE) architecture has become a fundamental building block in state-of-the-art large language models (LLMs), improving domain-specific expertise in LLMs and scaling model capacity without proportionally increasing their computational overhead. However, MoE inference often suffers from suboptimal GPU utilization, load imbalance, and elevated latency arising from multiple tokens waiting on the same experts for their computation which arises from sparsity of expert activation.

To address these challenges, we propose a \textit{dynamic} expert replication strategy that predicts which experts are likely to be overloaded and replicates them for upcoming batches of tokens. The replicated experts process batch tokens concurrently across layers, which leads to improved parallelism, shorter GPU idle time, and significantly faster inference. Experimental evaluations conducted on large-scale MoE models, including Switch-base-128 and Switch-base-256, demonstrate that our method achieves near-complete GPU utilization ($\sim$100\%), leading to upto 3x improvement in inference speed while preserving approximately 90-95\% of the performance of baseline architectures.
\end{abstract}

\maketitle

\section{Introduction}
Recent breakthroughs in machine learning, particularly in natural language processing, have primarily been enabled by exponential growth in computational resources, expansive training datasets, and increasingly larger models. However, training and deploying such massive models typically require extensive computational setups involving thousands of interconnected GPUs or accelerators running for prolonged periods, resulting in significant costs and energy consumption. Consequently, the machine learning community actively pursues more computationally efficient approaches, notably using sparse expert architectures.

The Mixture-of-Experts (MoE) model is an exemplary sparse architecture that partitions neural network parameters into distinct subsets called "experts," with each expert selectively handling specific input samples. MoE architectures significantly reduce computational overhead compared to traditional dense networks, which apply all parameters to every input by conditionally routing inputs to relevant experts. Prominent language models such as GPT-4, Mixtral, Llama 4 and more highlight this evolution. This conditional computation strategy, first introduced by Shazeer \cite{shazeer2017}, has delivered state-of-the-art results in various domains, including natural language modeling, translation, computer vision, speech recognition, and multimodal learning. Recent comprehensive studies have also validated the scalability and effectiveness of MoE configurations across diverse model sizes and expert counts \cite{moesurveypaper} \cite{millionmoe}.

\begin{figure}[htbp]
  \centering
  \includegraphics[width=\linewidth]{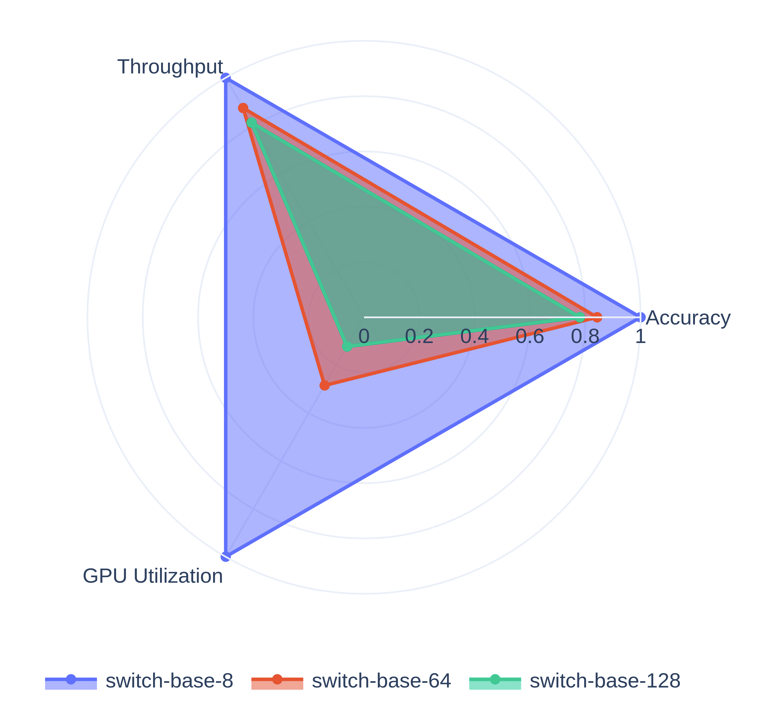}
  \caption{Issues in MoE-based SwitchTransformers\\(MultiRC SuperGlue dataset)}
  \label{fig:intro}
\end{figure}

Despite these benefits, MoE architectures face substantial challenges during inference, particularly in latency, inefficient memory usage, and suboptimal GPU utilization. In practice, MoE inference can be substantially slower—up to 15 times for language modeling and approximately three times slower for machine translation—than equivalent dense models. Existing optimization techniques like distillation and GPU kernel-level enhancements typically provide limited or narrowly focused improvements, often at the expense of model accuracy or general applicability. Various SwitchTransformers have been compared in the terms of GPU utilisation, throughput and accuracy in the figure \ref{fig:intro} for MultiRC dataset under SuperGlue when trained for the similar amount of time, values are normalized.

In this paper, we address these critical inefficiencies by exploring predictive prefetching and expert replication techniques inspired by concepts of multiple producers and multiple consumers (MPMC) from distributed systems. Leveraging these insights allows us to efficiently manage concurrent expert access and resource allocation in GPU environments. Additionally, we demonstrate through rigorous experiments that predicting expert selections for tokens is not as challenging as traditionally perceived. Our findings show that accurate estimation of expert requirements is feasible when fine-tuning and extensive pretraining occur concurrently with token prediction tasks. This approach substantially enhances inference speed, optimizes GPU utilization, and maintains model accuracy, representing a significant step toward practical and economically viable deployment of MoE-based large-scale language models. Our contributions are threefold:

\begin{enumerate}
    \item We introduce the highly efficient Simple Recurrent Unit (SRU) \cite{sru} to predict expert selections rapidly for each token, significantly speeding up inference.
    \item We propose using deep physical replication of experts, which is different for each expert based on prediction, ensuring that almost no token experiences waiting delays during batch processing, thus substantially increasing GPU utilization, throughput, and also reducing inference latency.
    \item We implement a practical strategy of capping expert replication at the GPU capacity, ensuring that replication is optimized within predefined GPU constraints to achieve near-optimal resource utilization.
\end{enumerate}

These innovations significantly enhance inference speed, optimize GPU utilization, and maintain model accuracy, representing a significant step toward the practical and economically viable deployment of MoE-based large-scale language models.

\section{Preliminaries}
The prevalent challenges in Mixture-of-Experts (MoE) frameworks arise predominantly from the inherent sparse activation of experts during inference. In large language models (LLMs) employing MoE architectures, this sparsity induces significant inefficiencies in key performance metrics: latency, GPU utilization, and throughput. Latency, defined as the overall inference time for a given test dataset, is adversely affected by inefficient all-to-all routing, where excessive inter-expert communication and the overhead incurred by routing via softmax contribute to additional delays. The softmax-based routing mechanism, while effective in handling probabilistic assignments, introduces computational complexity that does not scale favorably with the increasing number of experts and tokens, thereby further exacerbating latency issues.

\begin{figure}[htbp]
  \centering
  \includegraphics[width=0.6\linewidth]{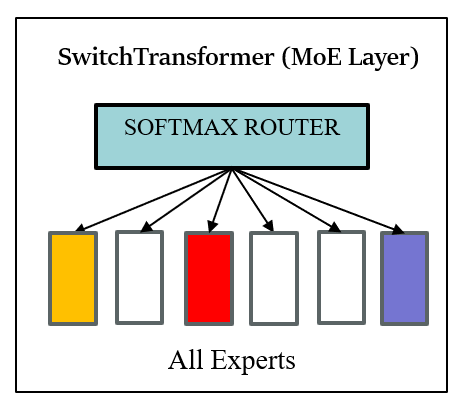}
  \caption{SwitchTransformers MoE Layer with all experts, active and inactive(white) in GPU}
  \label{fig:baselinemoe}
\end{figure}

Simultaneously, GPU utilization is suboptimal because, at any moment, only a limited subset of experts is active while the majority of GPU resources remain idle. This phenomenon is compounded by the fact that the dynamic routing process—involving both softmax computation and assignment of tokens—does not sufficiently leverage the parallel processing capabilities of modern GPUs. Moreover, token stalling is observed when the routing algorithm fails to distribute the computational load evenly, which directly suppresses throughput, measured as the number of samples processed per second. This study is further motivated by the observation that traditional routing approaches, including softmax-based mechanisms, contribute substantially to the observed inefficiencies. The computational burden of softmax routing, combined with all-to-all communications and the uneven load distribution across experts, necessitates novel architectural adaptations \cite{pregatedmoe} \cite{tutel} \cite{sidamoe}.

\paragraph{MoE Overhead}
The growing number of experts in MoE models introduces significant sparsity, which results in substantial under-utilization of GPU resources. As more experts are added, a larger proportion of GPU memory sits idle during each forward pass, leading to severe inefficiencies during inference. This sparsity not only wastes precious compute resources but also increases the overhead associated with expert selection and communication. In large-scale models, such as those featuring 128 or 256 experts, this overhead can account for over 72\% of the inference time, severely impacting overall performance as is shown in the introduction diagram \cite{sidamoe}.

\begin{figure}[htbp]
  \centering
  \includegraphics[width=0.6\linewidth]{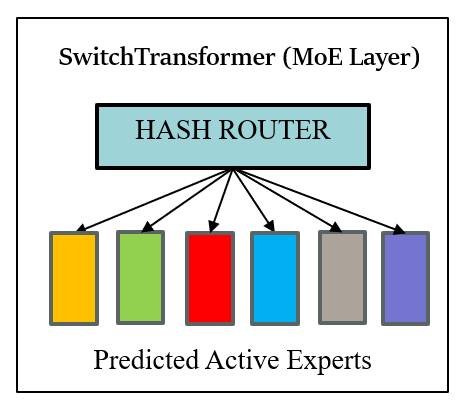}
  \caption{SiDA-MoE MoE Layer with only predicted active distinct experts in GPU}
  \label{fig:sidamoe}
\end{figure}

\begin{figure}[htbp]
  \centering
  \includegraphics[width=0.6\linewidth]{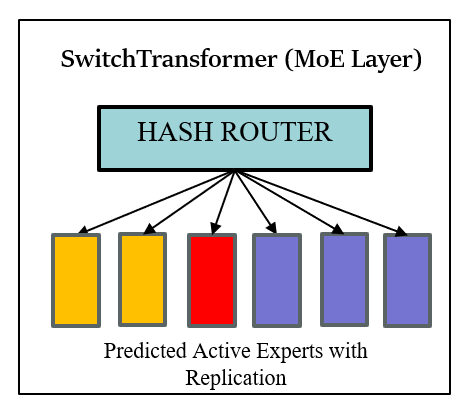}
  \caption{MoE-MPMC(Ours) MoE Layer with predicted active experts with their replicas in GPU}
  \label{fig:moempmc}
\end{figure}

\paragraph{Need for placing active experts in GPU with replication}
The selective and sparse activation of experts results in a highly imbalanced computational load distribution. For example, with a batch size of 64 and MoE layer comprising of 64 experts—where half of the tokens are routed to one expert and the remainder to another—only 2 out of 64 experts in the layer are heavily utilized, while the remaining experts remain largely idle. This imbalance critically undermines the system’s parallelization capacity and overall throughput. Tokens being routed to the same expert will end up waiting for the same physical experts, leading to queueing. It would be reasonable to put only the 2 active experts in GPU. Innovations like next-batch expert prediction approach have started to address these challenges; for instance, hash predictor in SiDA-MoE \cite{sidamoe} achieves nearly 99\% accuracy in aligning with the actual prediction routing mechanism of softmax. This demonstrates a strong potential to reduce these inefficiencies of GPU utilisation by placing only the active experts for the next in GPU. A promising strategy, suggested by our approach, to address these issues also involves preemptively predicting expert activation and subsequently replicating the predicted experts physically on device based on token demand. By forecasting which experts will be active prior to inference, it becomes feasible to replicate these experts—thereby increasing parallelization, throughput, and effective GPU utilization. In the given scenario, replicating the 2 active experts 32 times each would substantially alleviate the computational bottleneck, allowing each incoming token to be processed by a distinct physical expert replica and maximizing the effective use of GPU FLOPS. This method also does not change the existing baseline MoE model's computation graph. This is the main intuition behind our method.

The accompanying diagrams in this section illustrate the comparative MoE layers of the baseline scenario in SwitchTransformers which involve all experts, active or inactive, resident on the GPU (\ref{fig:baselinemoe}); the SiDA-MoE configuration limits GPU allocation to only those experts predicted to be active for the current batch(\ref{fig:sidamoe}); and the proposed approach; the proposed approach MoE-MPMC extends the strategy in SiDA-MoE by replicating the predicted experts physically in proportion to their anticipated token load (\ref{fig:moempmc}).

\section{Methodology}
\begin{figure*}[htbp]
  \centering
  \includegraphics[height=0.95\textheight,width=0.65\linewidth]{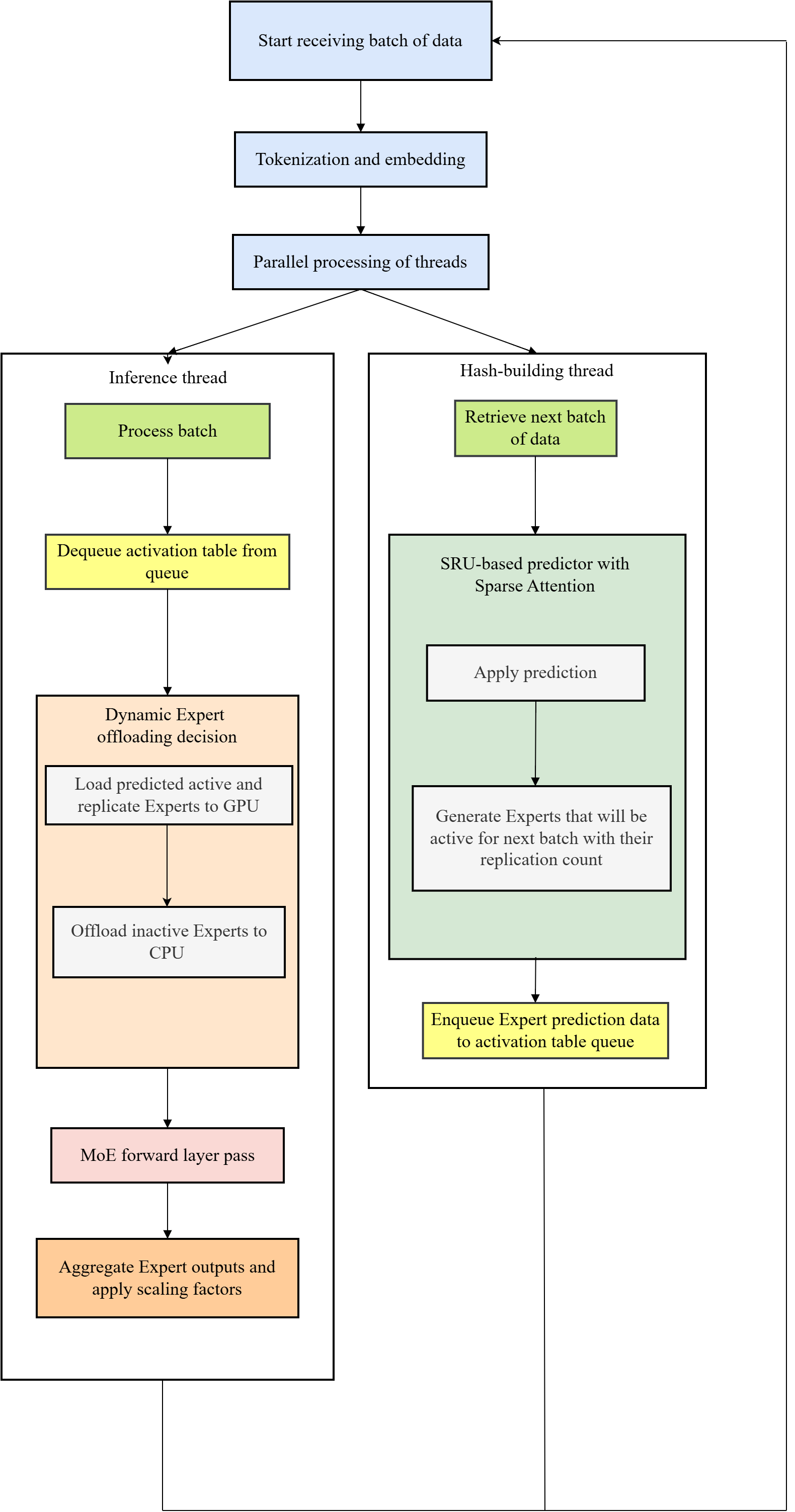}
  \caption{Workflow of Fast MoE Inference using MoE-MPMC.}
  \label{fig:workflow}
\end{figure*}

Our framework, named MoE-MPMC, processes input data using two concurrent threads: an inference thread and a hash-building thread. This design builds on the foundational ideas of SiDA-MoE (Sparsity-inspired Data-Aware Mixture-of-Experts) \cite{sidamoe}. The inference thread tokenizes incoming batches and executes core inference tasks. These tasks include dequeuing a specialized hash table from a queue, each hash table element briefly storing expert identifiers and their replication counts for the current batch. In this table, the key is a combination of the token and layer number, and the value comprises expert indices along with their replication counts. Only the experts predicted to be active are offloaded to the GPU with their deep copies. Subsequently, current batch inference is performed using a model similar to SwitchTransformers (top-1), where the MoE layer utilizes GPU-resident experts for enhanced efficiency. In parallel, the hash-building thread employs an SRU-based predictor with sparse attention to forecast the experts needed for the next batch, computing replication counts with a specialized hash algorithm and updating the hash table accordingly. The updated hash table is then enqueued back into the processing queue, and the entire process continues cyclically. The workflow below in (\ref{fig:workflow}) highlights this procedure, and detailed discussions of each component will follow in the subsequent sections.

\subsection{Inference thread}
The inference process initiates by retrieving a hash table from the queue for each input batch. This operation is performed concurrently with a dedicated parallel hash-building thread, which constructs the table that delineates the expert activation pattern for every Mixture-of-Experts (MoE) layer. By replicating the softmax routing with a more computationally efficient and parallelizable hashing mechanism, the system significantly reduces computational overhead. Following the retrieval of the hash table, the inference thread dynamically loads the required experts into GPU memory while simultaneously offloading unused experts to the main memory based on the prescribed activation pattern. This mechanism not only minimizes GPU memory usage but also reduces inference latency, ensuring that the inference thread remains optimally active while the hash-building thread continuously prepares the necessary activation maps.

The hash table further incorporates the details of the maximum replication counts for each expert predicted to be active in the current batch. These counts determine the maximum number of physical instances of an expert that should be deployed on the GPU. This replication strategy guarantees that when multiple tokens require the same expert, almost all token requests are processed in parallel by directing them to distinct physical copies of experts. It is important to note that parallel processing of tokens is maintained as long as the batch size remains within the capacity of the experts loaded onto the GPU. Detailed management of capacity constraints is addressed in subsequent sections. The dynamic, on-demand loading, offloading, and replication of experts—within established capacity limits—collectively contribute to a scalable and efficient inference process, even under scenarios of constrained GPU memory. The algorithm for this process is described in detail in (\ref{alg:fastmoe_inference}).

\begin{table}[ht]
\small
\centering
\caption{Glossary for algorithms}
\label{tab:glossary}
\begin{tabular}{p{0.18\linewidth} p{0.75\linewidth}}
\toprule
\textbf{Term} & \textbf{Definition} \\
\midrule
$\mathbf{X}_i$ 
& The \(i\)-th input batch of token embeddings processed by the model. \\[6pt]

$\mathbf{H}_i$
& The hash table for the \(i\)-th batch, storing the expert assignments for each token in each MoE layer. \\[6pt]

$\mathbf{h}_i$
& The predictor hash table (trained during finetuning) that predicts expert assignments for each token in each MoE layer. \\[6pt]

$E_i$
& The set of all expert indices offloaded to CPU memory for the \(i\)-th batch. \\[6pt]

$l$
& MoE layer index. \\[6pt]

$s$
& A token within the current batch. \\[6pt]

$k$
& The selected expert index for a particular token in a particular MoE layer. \\[6pt]

$M$
& The base MoE model (or neural network) comprising multiple experts. \\[6pt]

$M_{\text{MoE-MPMC}}$
& A specialized MoE model supporting dynamic replication/assignment of experts for faster parallel inference. \\
\bottomrule
\end{tabular}
\end{table}

\begin{algorithm}[htbp]
\caption{Fast MoE Inference using MoE-MPMC (Inference Thread)}
\begin{algorithmic}[1]
\FOR{each batch $\mathbf{X}_i$}
    \STATE Dequeue $\mathbf{H}_i$\\
    \FOR{each MoE layer $l$}
        \STATE Set of experts $E_i$ = all expert indices in layer l
        \FOR{each token $s$}
            \STATE Check Expert $\mathbf{H}_i[l][s]$
            \IF{not on GPU}
                \STATE add single replica
                \STATE remove expert index from $E_i$
            \ELSIF{present on GPU}
                \STATE remove current expert index from $E_i$
                \STATE append physical replica of Expert $\mathbf{H}_i[l][s]$ with $new\ index$\\
                $\mathbf{H}_i[l][s]$ = $\text{Expert}_{new\ index}$ 
                \STATE (Physical replicas added within cap)
            \ENDIF
        \ENDFOR\\
        Make sure to move all experts in $E_i$ to CPU
    \ENDFOR\\
    \STATE $M_{\text{MoE-MPMC}} \leftarrow M$
    \STATE output = $M_{\text{MoE-MPMC}}(\mathbf{X}_i)$
\ENDFOR
\end{algorithmic}
\label{alg:fastmoe_inference}
\end{algorithm}

\subsection{Hash-building thread}
The hash building thread runs in the background, fetches the next batch of tokens (while the current batch is running in inference), and predicts for each token in each layer, the experts that it needs. The predicting hash table is constructed during finetuning and used during inference. SRU \cite{sru} is used as a predictor in the hash table during the finetuning phase. Sparsemax is used here to understand which experts would be activated. The detailed algorithm with its flow is given in the algorithm \ref{alg:fastmoe_hash}.

\begin{algorithm}[htbp]
\caption{Fast MoE Daemon Hash building using MoE-MPMC (Hash Building Thread), same as SiDA-MoE}
\begin{algorithmic}[1]
\FOR{each batch $\mathbf{X}_i$}
    \FOR{each MoE layer $l$}
        \FOR{each token $s$}
            \STATE Expert $k \leftarrow h_l(\mathbf{X}_i[s])$
            \STATE $\mathbf{H}_i[l][s] \leftarrow$ Expert $k$
        \ENDFOR
    \ENDFOR
    \STATE Enqueue $\mathbf{H}_i$
\ENDFOR
\end{algorithmic}
\label{alg:fastmoe_hash}
\end{algorithm}

Our approach advances the hash building process by replacing the conventional LSTM \cite{lstm} with a Simple Recurring Unit (SRU) \cite{sru} to predict expert activations. SRUs, compared to LSTMs, offer several computational benefits, including faster training and inference speeds due to their simplified gating mechanisms and fewer trainable parameters. The simpler structure of SRUs reduces computational overhead and enables rapid inference, making them highly suitable for environments where low latency and high throughput are critical. A detailed exploration of SRU's benefits over LSTM, including experimental results demonstrating these advantages, is provided in later sections of this paper.

To ensure optimal resource utilization without exceeding hardware constraints, we strategically cap expert replication according to GPU memory capacity. This preconfigured cap prevents excessive replication, maintaining the balance between optimal GPU utilization and resource availability. By setting appropriate replication limits, we ensure sustained high GPU utilization, stable system performance, and effective load balancing.

Collectively, our integration of SRU-based predictive prefetching and expert replication significantly enhances the SiDA-MoE framework. This combined methodology delivers substantial efficiency improvements, notably reduced inference latency and optimized GPU resource utilization, providing clear advantages over traditional MoE inference methods.

\subsection{Simple Recurrent Unit(SRU) for expert prediction}
The Simple Recurrent Unit (SRU) substantially reduces the inter-dependencies brought by LSTM, thereby facilitating enhanced parallel computation and lowering overall training time without compromising model performance.

The SRU architecture is defined by a streamlined set of operations. For an input \(x_t\) at time step \(t\), the SRU computes its hidden representations as follows:
\[
\begin{aligned}
x'_t &= W x_t, \\
f_t &= \sigma(W_f x_t + b_f), \\
r_t &= \sigma(W_r x_t + b_r), \\
c_t &= f_t \odot c_{t-1} + (1 - f_t) \odot x'_t, \\
h_t &= r_t \odot g(c_t) + (1 - r_t) \odot x_t,
\end{aligned}
\]
where \(W\), \(W_f\), and \(W_r\) denote the trainable weight matrices, \(b_f\) and \(b_r\) are the corresponding bias vectors, \(\sigma(\cdot)\) is the sigmoid activation function, \(g(\cdot)\) is a non-linear activation (typically the hyperbolic tangent), and \(\odot\) represents element-wise multiplication. The simplified computation model of the SRU enables more efficient state propagation and training dynamics.

In practical implementations, it is advantageous to re-engineer recurrent modules originally based on LSTM architectures to accommodate SRU cells. Careful initialization of SRU parameters is essential to prevent early saturation and ensure robust training stability. We use a 10-layer SRU for our predictions. By adhering to the SRU formulation for state propagation, the resulting architecture leverages improved parallelism and reduced computational cost. Empirical evaluations on standard sequence modeling benchmarks have shown that replacing LSTM with SRU maintains competitive performance while achieving up to a 30\% reduction in training time for MoE-MPMC.
SparseMax is employed to produce a sparse distribution over attention weights, assigning zero probability to less-relevant tokens. This allows the model to focus on the most critical embeddings while reducing overhead, improving prediction efficiency.

\subsubsection{Capacity capping}

The introduction of adding physical replicas may, in some cases, cause overflow in the GPU depending on the batch of tokens. To avoid this, it is necessary that the count of experts is capped to the count allowed in the GPU. This is done by manually setting the expert limit for each expert by going over the hash table once more to assign maximum number of replicas per expert. Each distinct expert gets a count of atleast 1. We then use the heuristic of dividing the remaining experts space in GPU by the count of remaining predicted replicated experts to be added and those capacity limits are increased accordingly.

Through extensive experimentation, it has been observed that the incorporation of expert replication contributes to a balanced and efficient inference process as seen in tbe experiments section.

\section{Experiments}
In our experiments, we utilized a high-performance computing (HPC) cluster and used SLURM for  configuring and running the fine-tuning tasks. The experimental setup included a single NVIDIA A100 GPU with 80GB of GPU memory and node with 256GB of main memory powered by an Intel® Xeon® Gold 6140 CPU running at 2.30GHz. All evaluations were conducted using the SwitchTransformers base model for 8, 64 and 128 experts. For the SST2 GLUE dataset \cite{sst2}, the model was fine-tuned using the top-1 configuration, whereas for the MRPC GLUE dataset \cite{mrpc}, a top-3 configuration was employed.

\begin{figure*}[htbp]
  \centering
  \includegraphics[width=\linewidth]{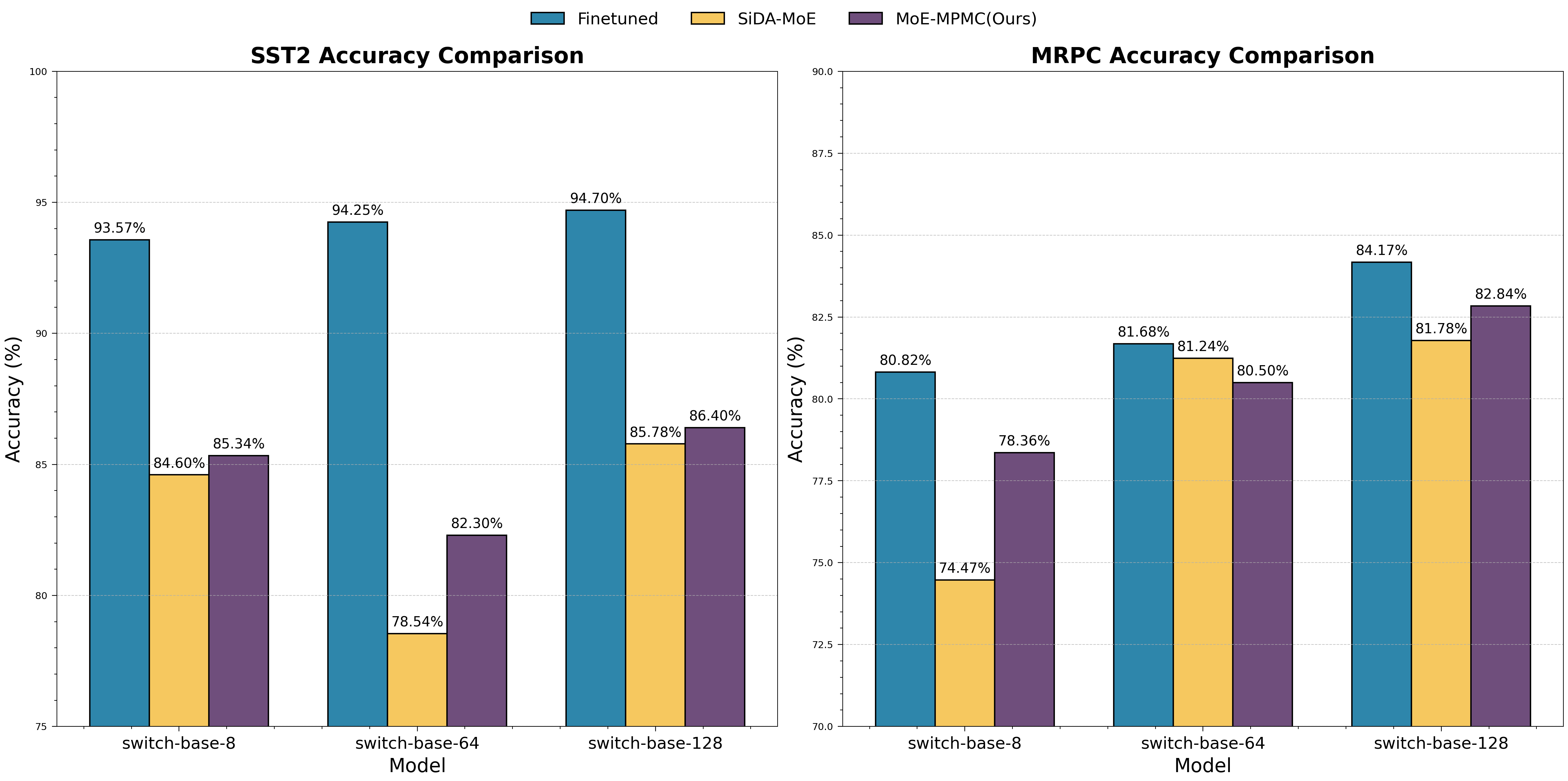}
  \caption{Comparison of accuracies between SwitchTransformers, SiDA-MoE and MoE-MPMC.}
  \label{fig:accuracy}
\end{figure*}

\begin{figure*}[htbp]
  \centering
  \includegraphics[width=\linewidth]{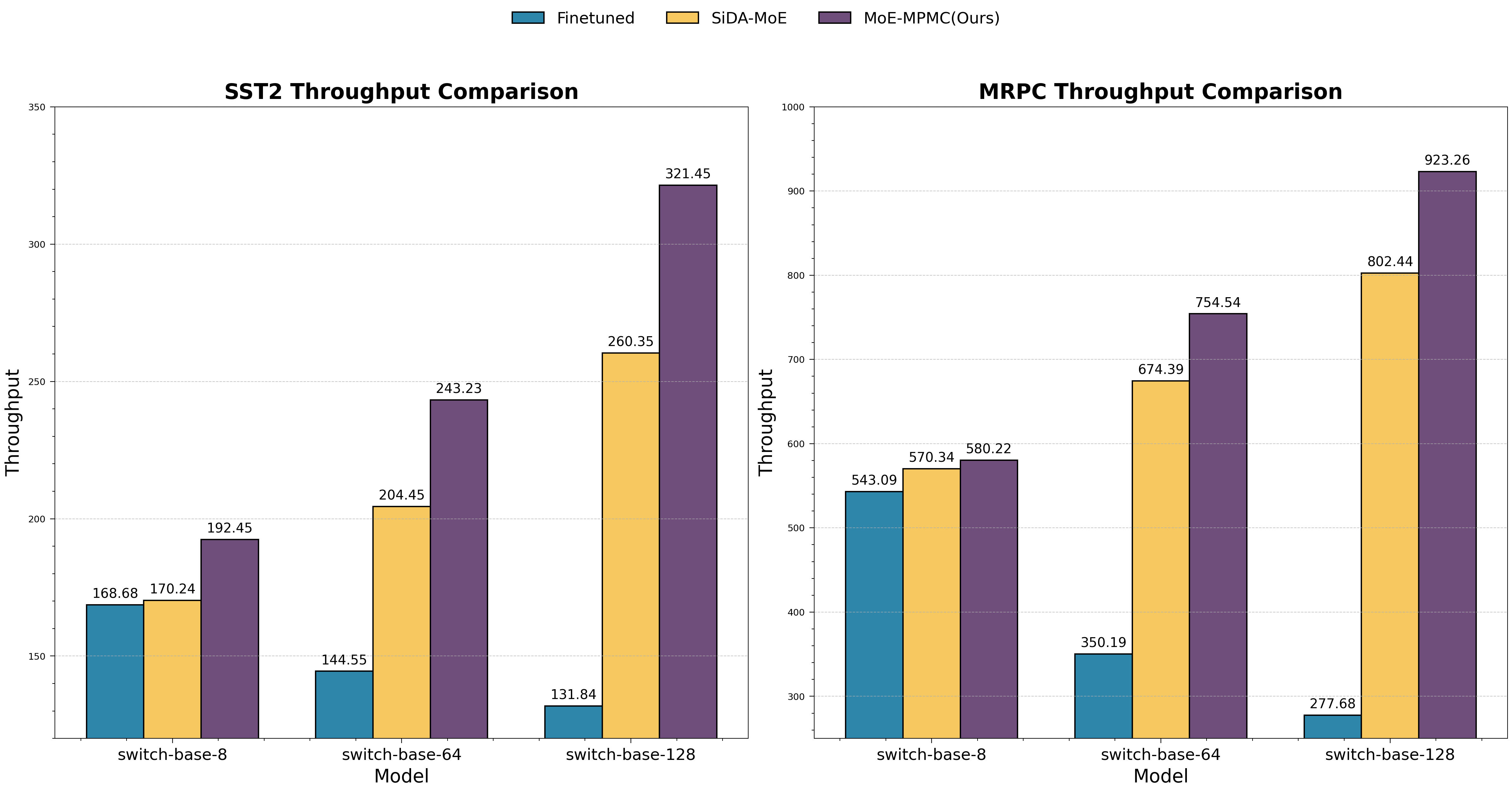}
  \caption{Comparison of throughput between SwitchTransformers, SiDA-MoE and MoE-MPMC.}
  \label{fig:throughput}
\end{figure*}

\begin{figure*}[htbp]
  \centering
  \includegraphics[width=\linewidth]{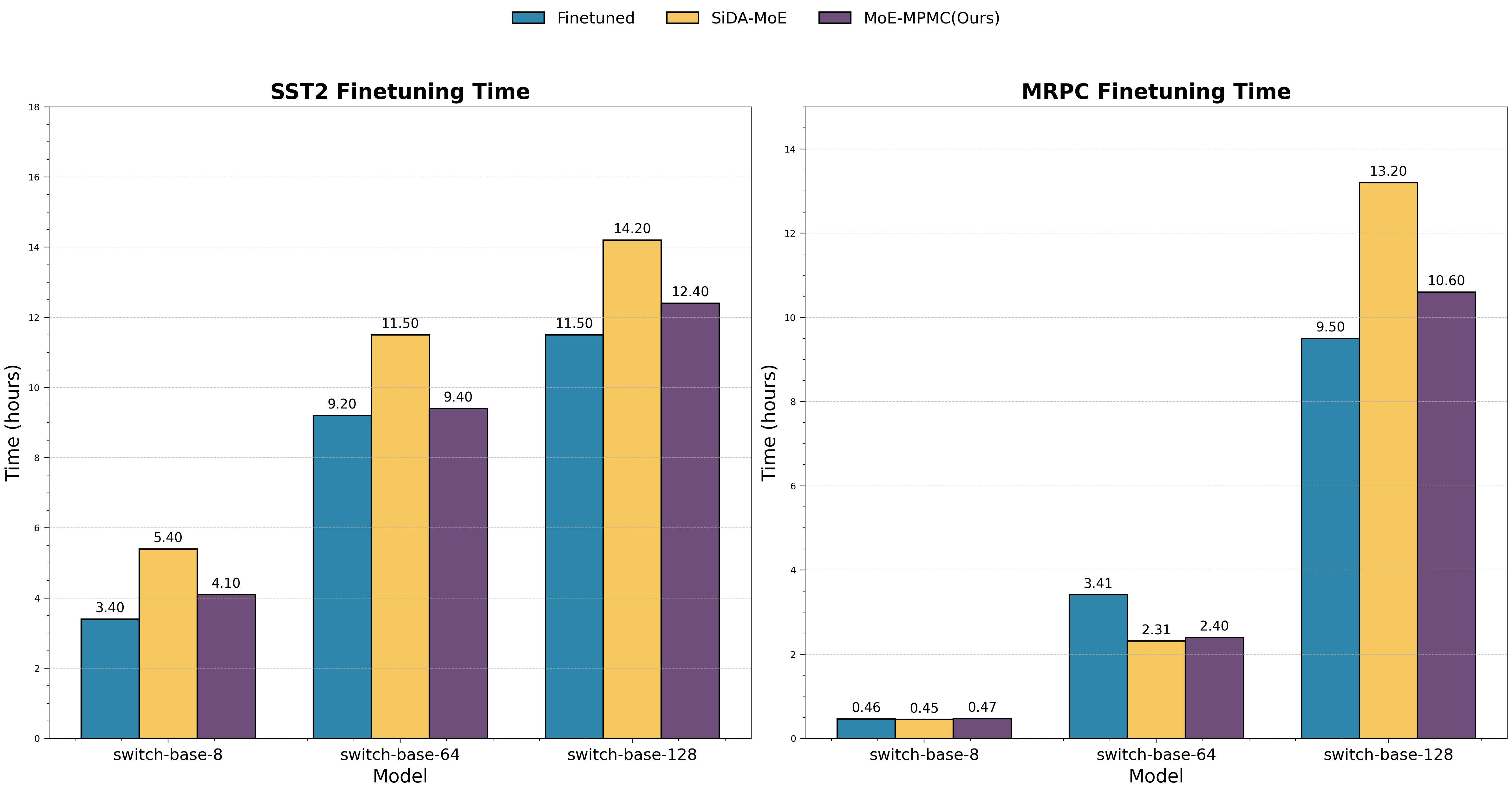}
  \caption{Comparison of finetuning time between SwitchTransformers, SiDA-MoE and MoE-MPMC.}
  \label{fig:finetune}
\end{figure*}

\begin{table}
  \caption{GPU Utilisation for Different Models}
  \label{tab:gpu_utilisation}
  \begin{tabular}{lccc}
    \toprule
    Model & Finetuned & SiDA-MoE & MoE-MPMC \\
    \midrule
    switch-base-8   & 7.5\%   & 89\%   & 96\% \\
    switch-base-64  & 3.4\%   & 79\%   & 89\% \\
    switch-base-128 & 1.3\%   & 63\%   & 82\% \\
    \bottomrule
  \end{tabular}
\end{table}

To assess model performance and efficiency, we employed comparisons with baseline SwitchTransformers, SiDA-MoE and MoE-MPMC(ours). We captured detailed measurements of latency, GPU utilization, fine-tuning accuracy, overall accuracy, and throughput. Capacity limits of replication counts have not been setup for the experiments as it was observed the sparsity was high and did not require capacity capping. The experimental results clearly demonstrate the superiority of the MoE-MPMC approach, revealing significant improvements across multiple performance indicators. The count of epochs is kept at 20000, learning rate 0.001, batch-size 64 have been kept the same across the models for the different experiments.  

In the table (\ref{tab:gpu_utilisation}), we observe the general trend of lowering GPU utilisation as the count of experts increases. The gpu utilisation is maintained very well by both the SiDA-MoE and MoE-MPMC models. MoE-MPMC has highest effective gpu utilisation amongst the three. We can also the see results of accuracy for the three models across switch-base-8, switch-base-64 and switch-base-128 (\ref{fig:accuracy}). Accuracy is maintained better when using MoE-MPMC in comparison with SiDA-MoE primarily because of better prediction by SRU over LSTM . Throughput also shows improvement in MoE-MPMC because of multiple replicas of experts. Throughput reaches higher values as the number of experts increases (\ref{fig:throughput}) . The interesting thing to note is that the training dynamics of the larger expert models makes the convergence take longer time. The finetuning time for the various models are also provided (\ref{fig:finetune}). Learning the entire batch level expert requirement does cause additional computation flops with the baseline and though the numbers are more than the baseline, they are not that high as compared to SiDA-MoE.

In light of these results, we can see how a dynamic replication strategy not only harnesses the potential of MoE-based architectures but also addresses the typical pitfalls such as idle GPU memory and load imbalances. By enabling multiple concurrent expert replicas and efficiently handling sparse token allocations, MoE-MPMC consistently matches or surpasses the performance of previous approaches like SwitchTransformers and SiDA-MoE. This approach offers a robust alternative for scenarios where large-scale MoE models must be both accurate and operationally efficient, serving as a promising direction for future work and real-world deployment.

\section{Related works}
Mixture-of-Experts (MoE) architectures have emerged as a transformative approach for scaling large language models (LLMs) by leveraging sparse computation through selective expert activation \cite{fedus2022reviewsparseexpertmodels}. Introduced by Shazeer \cite{shazeer2017}, MoEs enable massive parameter scaling, upto billions and more, while maintaining relatively constant inference costs. Comparative studies, such as those on wider FFN vs. MoE Transformers \cite{ffnvstrans}, critically examine trade-offs between simply widening feedforward networks and employing MoEs, underscoring MoEs’ superior scalability despite the increased complexity in routing and training stability. However, there exist challenges that hinder real-world deployment at scale. Most of these issues were related to balancing load, effective GPU utilization and managing sparsity without incurring communication overhead, accuracy loss or latency overhead. To overcome these issues, recent research has pursued both architectural and system-level optimizations for MoE-based model inference. Comprehensive survey papers \cite{moesurveypaper} have outlined many previous works on MoE inference optimizations, many of which have synthesized theoretical and practical advancements in sparse expert-based models, highlighting the critical importance of efficient MoE design for inference.

\subsection{Model-Level Improvements}
At the model level, research efforts target three major dimensions: algorithmic enhancement, architecture design and model compression. In terms of algorithmic improvement, approaches like AdaptMoE \cite{adapmoe} and DynMoE \cite{dynamoe} have explored dynamic gating schemes, thereby reducing computational overhead without compromising model quality. AdapMoE is an algorithm-system co-design framework for Mixture-of-Experts inference that adaptively adjusts expert gating, prefetching, and caching based on sensitivity analysis for improved inference. DynMOE is an auto-tuning approach for Transformer models that dynamically adjusts the number of activated experts per token to optimize efficiency and performance. For architecture design, frameworks such as MoH \cite{moh}, JetMoE \cite{jetmoe} integrate the MoE concept into both attention and feed-forward layers, leveraging sparse computations to scale efficiently while preserving performance on a variety of downstream tasks. Lastly, model compression leverages established pruning, quantization, and distillation techniques like by CMoE \cite{cmoe} to tailor MoE models to resource-constrained devices. These compression strategies frequently employ expert pruning or low bit quantization to reduce memory footprints and reduce communication overhead, while knowledge distillation-based approaches, similar to DeepSpeed-MoE \cite{deepspeedmoe}, effectively transfer sparse representations into more compact dense or semi-sparse student models. 

\subsection{System and Hardware-Level Optimizations}
Beyond architectural refinements, the system level addresses the distribution and offloading complexities inherent to large-scale MoE deployments. Solutions—such as Gshard \cite{gshard}, FastMoE \cite{fastmoe}, and Tutel \cite{tutel}, which introduced reduce-scatter, coordinate how experts communicate across multiple GPUs or nodes, reducing the bottlenecks caused by load imbalance and excessive communication. Building on this foundation, frameworks like Alpa \cite{alpa} and FlexMoE \cite{flexmoe} blend data, tensor, and pipeline parallelismto maximize throughput. Concurrently, expert offloading techniques aim to overcome memory constraints, particularly on edge devices. Works such as HOBBIT \cite{hobbit}, AdapMoE \cite{adapmoe} selectively prefetch high-traffic experts from CPU into GPU memory based on predicted activation patterns, thereby hiding load latencies and cutting down the overhead of frequently accessing off-device storage. Pre-gated MoE \cite{pregatedmoe} introduces a preemptive gating mechanism that decouples expert selection from execution to overlap CPU–GPU transfers, dramatically reducing latency and memory overhead for scalable MoE inference. By uniting model, system, and hardware-level innovations, these efforts collectively move MoE models closer to practical, large-scale deployment with stronger performance and resource efficiency.

These techniques were fundamentally reactive; there have been many developments to anticipate future demands or dynamically adjust to evolving system conditions to enhance their applicability in real-world systems. On the deployment side, systems such as SEER-MoE \cite{seermoe} and EdgeMoE \cite{edgemoe} have achieved notable improvements in inference throughput through optimized scheduling of memory movements and hardware-aware execution pipelines. For instance, EfficientMoE \cite{zadouri2024pushing} introduces an adaptive load balancing strategy that dynamically distributes computation across experts, significantly enhancing both training scalability and convergence speed. Complementing this, APTMoE \cite{aptmoe} proposes affinity-aware pipeline tuning for GPUs to reduce inter-node communication by aligning expert scheduling with the underlying hardware topology. Moreover, the Mixture of a Million Experts \cite{millionmoe} framework pushes MoE scalability by employing a highly sparse gating mechanism with millions of experts that activates only a few relevant experts per input, thereby preserving computational efficiency and reducing load per expert. Techniques such as SiDA-MoE predict expert activation patterns for inference optimization \cite{sidamoe}, providing an orthogonal perspective to MoE inference. It uses the concept of expert prediction for the entire next batch and runs the inference and prediction in two parallel threads. Our work, MoE-MPMC, introduces a data-aware, look-ahead strategy that anticipates expert activation and replication patterns, building upon the foundational work of SiDA-MoE. 

\section{Conclusion}
Our approach differs from reactive methods by proactively prefetching and replicating the necessary experts. This strategy mitigates memory access bottlenecks and improves throughput, all while preserving the underlying MoE graph computation as demonstrated in our experiments. We propose using SRU as a promising model for expert prediction in this domain due to its high degree of parallelization and superior performance compared to the traditional LSTM approach introduced in SiDA-MoE. The inherent correlations and token routing between different MoE layers further incentivize the use of expert prediction. In summary, our method maintains the advantages of model sparsity while significantly accelerating expert access and thereby inference—a critical advancement for next-generation MoE systems.

\bibliographystyle{ACM-Reference-Format}
\bibliography{sample-base}

\end{document}